%% file: main.tex
\documentclass[11pt,a4paper]{article}
\usepackage[hyperref]{conll-2019}
\usepackage{times}
\usepackage{latexsym}
\usepackage{MnSymbol}
\usepackage{booktabs}
\usepackage{paralist}
\usepackage{amsmath}
\usepackage{wrapfig}
\usepackage{graphicx}
\usepackage{pgfplots}
\usepackage{subcaption}
\usepgfplotslibrary{groupplots}
\usetikzlibrary{matrix}
\pgfplotsset{compat=newest}
\usepackage{enumitem}
\graphicspath{ {./resources/} }
\usepackage[ruled, linesnumbered,noend]{algorithm2e}
\usepackage{url}
\DeclareMathOperator*{\argmax}{argmax}

\newcommand\comment[1]{}
\newcommand\sS{\mathcal{S}}
\newcommand\labset{\sS_\text{lab}}
\newcommand\unlabset{\sS_\text{unlab}}
\newcommand\evalset{\sS_\text{eval}}

\newcommand\sD{\mathcal{D}}
\newcommand\sC{\mathcal{C}}
\newcommand\bestoracle{\textsc{BaseOracle}}
\newcommand\random{\textsc{Random}}
\newcommand\longest{\textsc{Longest}}
\newcommand\uncertainty{\textsc{Uncertainty}}

\aclfinalcopy

\title{On the Limits of Learning to Actively Learn Semantic Representations}

\author{Omri Koshorek$^{1}$ ~~~  Gabriel Stanovsky$^{2,4}$ ~~~ Yichu Zhou$^{3}$ \AND  Vivek Srikumar$^{3}$ ~~~ Jonathan Berant$^{1,2}$
   \\\\
    $^1$Tel-Aviv University, $^2$Allen Institute for AI \\
    $^3$The University of Utah, $^4$University of Washington \\
    \texttt{\{omri.koshorek,joberant\}@cs.tau.ac.il} \\ \texttt{gabis@allenai.org}, 
    \texttt{\{flyaway,svivek\}@cs.utah.edu}
}
\comment{
\author{Omri Koshorek \\
  Tel Aviv University \\
  \And
  Gabriel Stanovsky \\
  University of Washington \\
  Allen Institute for AI \\
  \AND
  Yichu Zhou \\
  The University of Utah \\
  \And
  Vivek Srikumar \\
  The University of Utah \\
  \And
  Jonathan Berant \\
  Tel Aviv University \\
  Allen Institute for AI \\
   \\
}}

\date{}

\begin{document}
\maketitle
\input{00_abstract}
\input{01_intro}
\input{02_actlearn}

\input{03_oracle}

\input{04_qasrl_schema}
\input{05_basic_exps}

\input{06_extended_exsps}
\input{07_analysis}
\input{08_related}
\input{09_conclusions}
\input{11_ackno}
\bibliography{cites}
\bibliographystyle{acl_natbib}

\end{document}

%% file: 00_abstract.tex
\begin{abstract}
One of the goals of natural language understanding is to develop models that map sentences into meaning representations. 
However, training such models requires expensive annotation of complex structures, which hinders their adoption.
Learning to actively-learn (LTAL) is a recent paradigm for reducing the amount of labeled data by learning a policy that selects which samples should be labeled.
In this work, we examine LTAL for learning semantic representations, such as QA-SRL. We show that even an \emph{oracle} policy that is allowed to pick examples that maximize performance on the test set (and constitutes an upper bound on the potential of LTAL), 
does not substantially improve performance compared to a \emph{random} policy.
We investigate factors that could explain this finding and show that a distinguishing characteristic of successful \comment{versus unsuccessful} applications of LTAL is the interaction between optimization and the oracle policy selection process.
In successful applications of LTAL, the examples selected by the oracle policy do not substantially depend on the optimization procedure, while in our setup 
the stochastic nature of
optimization 
strongly affects the examples selected by the oracle.
We conclude that the current applicability of LTAL for 
improving data efficiency in
learning semantic meaning representations is limited.
\end{abstract}


%% file: 01_intro.tex
\section{Introduction}
The task of mapping a natural language sentence into
a \emph{semantic representation}, that is, 
a structure that represents its meaning, is one of the core goals of natural language processing. This goal has led to the creation of many general-purpose formalisms for representing the structure of language, such as semantic role labeling (SRL;  \citealp{palmer2005proposition}), semantic dependencies (SDP; \citealp{oepen2014semeval}), abstract meaning representation (AMR;  \citealp{banarescu2013abstract}), universal conceptual  cognitive  annotation (UCCA; \citealp{abend2013universal}), question-answer driven SRL (QA-SRL; \citealp{he2015question}), and universal dependencies \cite{nivre2016universal}, 
as well as domain-specific semantic representations for particular users in fields such as biology  \cite{kim2009overview,nedellec2013overview,berant2014modeling} and
material science \cite{Mysore2017AutomaticallyEA,Kim2019InorganicMS}.

Currently, the dominant paradigm for building models that predict such representations is supervised learning, which requires
annotating thousands of sentences with their correct structured representation, usually by experts.
This arduous data collection is the main bottleneck for building parsers for different users in new domains. 

Past work has proposed directions for accelerating data collection and improving data efficiency through multi-task learning across different representations \cite{Stanovsky2018EMNLP,hershcovich2018multitask}, or having non-experts annotate sentences in natural language~\cite{he2015question,he2016human}. 
One of the classic and natural solutions for reducing annotation costs is to use \emph{active learning}, an iterative procedure for selecting unlabeled examples which are most likely to improve the performance of a model, and annotating them \cite{settles2009active}. 
 
Recently, learning to actively-learn (LTAL) has been proposed~\cite{fang2017learning,bachman2017learning,liu2018learning}, where the procedure for selecting unlabeled examples is \emph{trained} using methods from reinforcement and imitation learning. In recent work by \newcite{liu2018learning},  given a labeled dataset from some domain, active learning is simulated on this dataset, and a policy is trained to iteratively select the subset of examples that maximizes performance on a development set.
Then, this policy is used on a target domain to select unlabeled examples for annotation. If the learned policy generalizes well,
we can reduce the cost of learning semantic representations.
\newcite{liu2018learning} and \newcite{vu2019learning} have shown that such learned policies significantly reduce annotation costs on both text classification and named entity recognition (NER).  

In this paper, we examine the potential of LTAL for learning a semantic representation such as QA-SRL. We propose an \emph{oracle setup} that can be considered as an upper bound to what can be achieved with a learned policy. Specifically, we use an \emph{oracle policy} that is allowed to always pick a subset of examples that maximizes its target metric on a development set, which has the same distribution as the \emph{test set}. Surprisingly, we find that even this powerful oracle policy does not substantially improve performance compared to a policy that randomly selects unlabeled examples on two semantic tasks: QA-SRL span (argument) detection and QA-SRL question (role) generation.

To elucidate this surprising finding, we perform a thorough analysis, investigating various factors that could negatively affect the oracle policy selection process. We examine possible explanatory factors including: (a) the search strategy in the unlabeled data space (b) the procedure for training the QA-SRL model (c) the architecture of the model and (d) the greedy nature of the selection procedure. We find that for all factors, it is challenging to get consistent gains with an oracle policy over a random policy.

To further our understanding, we replicate the experiments of \newcite{liu2018learning} on NER, and compare the properties of a successful oracle policy in NER to the less successful case of QA-SRL. 
We find that optimization stochasticity negatively affects the process of sample selection in QA-SRL;
different random seeds for the optimizer result in different selected samples.
\comment{given different random initializations of the QA-SRL model at each iteration} We propose a measure for quantifying this effect, which can be used to assess the potential of LTAL in new setups. 


To conclude, in this work, we conduct a thorough empirical investigation of LTAL for learning a semantic representation, and find that it is difficult to substantially improve data efficiency compared to standard supervised learning. 
Thus, other approaches should be explored for the important goal of reducing annotation costs in building such models. Code for reproducing our experiments is available at \url{https://github.com/koomri/LTAL_SR/}.

%% file: 02_actlearn.tex
\section{Learning to Actively Learn}
\label{sec:ltal}

Classic pool-based active learning ~\cite{settles2009active} assumes access to a small labeled dataset $\labset$ and a large pool of unlabeled examples $\unlabset$ for a target task. In each iteration, a heuristic is used to select $L$ unlabeled examples, which are sent to annotation and added to $\labset$. An example heuristic  is \emph{uncertainty sampling} \cite{Lewis1994ASA}, which at each iteration chooses examples that the current model is the least confident about.

\begin{figure}[t]
   \includegraphics[width=\columnwidth,keepaspectratio]{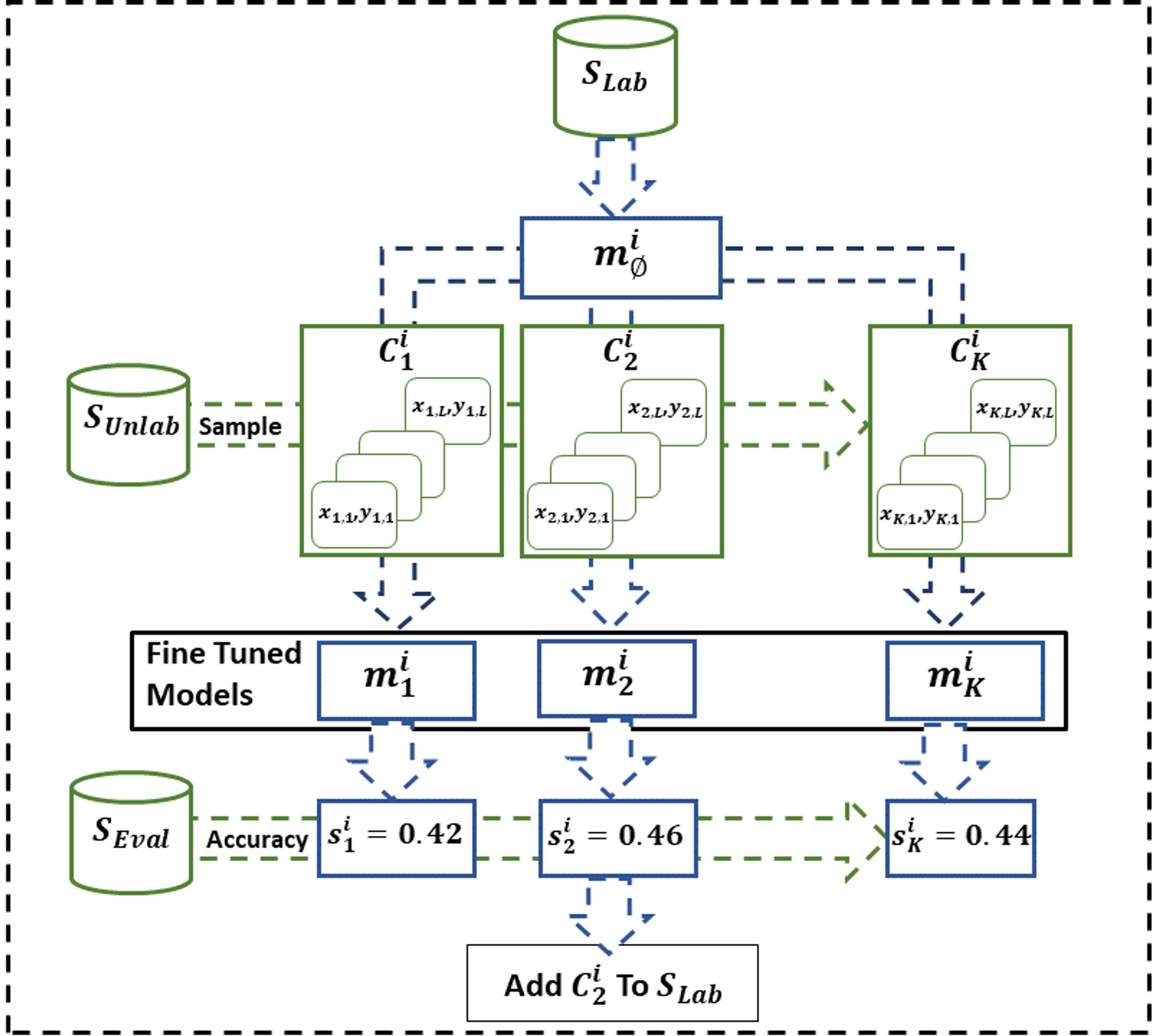}
\caption{A single iteration of LTAL, where examples are sampled from $\unlabset$, trained with examples in $\labset$, and performance on $\evalset$ is used to select examples to annotate. See \S\ref{sec:ltal} for details.}
\centering
\label{LTAL-fig}
\end{figure}

LTAL proposes to replace the heuristic with a learned policy $\pi_\theta$, parameterized by $\theta$. At \emph{training time}, the policy is trained by simulating active learning on a labeled dataset and generating training data from the simulation. At \emph{test time}, the policy is applied to select examples in a new domain.
Figure~\ref{LTAL-fig} and Algorithm~\ref{alg:algo-oracle}  describe this data collection procedure, on which we build our \emph{oracle policy} (\S\ref{sec:oracle}).

In LTAL, we assume a labeled dataset $\sD$ which is partitioned into three disjoint sets: a small labeled set $\labset$, a large set $\unlabset$ that will be treated as unlabeled, and an evaluation set $\evalset$ that will be used to estimate the quality of models. Then, active learning is simulated for $B$ iterations. In each iteration $i$, a model $m^i_\phi$, parameterized by $\phi$, is first trained on the labeled dataset.
Then, $K$ subsets $\{\sC_j\}_{j=1}^K$ are randomly sampled from $\unlabset$, and the model $m^i_\phi$ is fine-tuned on each candidate set, producing $K$ models $\{m^i_{\phi_j}\}_{j=1}^K$. The performance of each model is evaluated on $\evalset$, yielding the scores $\{s(\sC_j)\}_{j=1}^K$.
Let the candidate set with highest accuracy be $\sC^i_t$. We can create training examples for $\pi_\theta$, where $(\labset, \unlabset, m^i_\phi, \{s(\sC_j)\}_{j=1}^K)$ are the inputs and $\sC^i_t$ is the label. Then $\sC^i_t$ is moved from $\unlabset$ to $\labset$. 

Simulating active learning is a computationally expensive procedure. In each iteration we need to train $K$ models over $\labset \cup \sC_j$. However, a trained network can potentially lead to a policy that is better than standard active learning heuristics.

\input{10_algo_oracle}

%% file: 10_algo_oracle.tex
\SetKwInput{KwInput}{Input}                
\SetKwInput{KwOutput}{Output}   

\begin{algorithm}[t]
{\small
 
\caption{Simulating active learning}
\label{alg:algo-oracle}
\SetAlgoLined
\KwInput{$\labset, \unlabset, \evalset$ }
  
  \For{$i \in \{1 \dots B\}$}{

  $m^i_{\phi} \leftarrow \textnormal{Train}(\labset$)\label{line:train}

    $\sC^i_1,\dots,\sC^i_K = \textnormal{SampleCandidates}(\unlabset)$
  
  \For{$j \in \{1, \dots, K \}$}{ \label{line:for}
  
      
      $m^i_{\phi_j} \leftarrow \textnormal{FineTune}(m^i_\phi, \labset \bigcup \sC^i_j)$ \label{line:finetune}
      
      $s^i_j \leftarrow \textnormal{Accuracy}(m^i_{\phi_j}, \evalset)$ \label{line:score}
  }
  
  $\displaystyle t  \leftarrow \argmax_{j \in \{1, \dots ,K\}} s^i_j$
  
  $\textnormal{CreateTrainEx}((\labset, \unlabset, m^i_{\phi}, \{\sC^i_j\}_{j=1}^K),\sC^i_t)$ \label{line:create_train}
  
  $\labset \leftarrow \labset \bigcup 
  \sC^i_t$,
  $\unlabset \leftarrow \unlabset \setminus
  \sC^i_t$
 }
 \Return $\labset$
 }
\end{algorithm}

%% file: 03_oracle.tex
\section{An Oracle Active Learning Policy}
\label{sec:oracle}

Our goal is to examine the potential of LTAL for learning a semantic representation such as QA-SRL.
Towards this goal, we investigate an \emph{oracle policy} that should be an upper bound for what can be achieved with a learned policy $\pi_\theta$.

The oracle policy is allowed to use Algorithm~\ref{alg:algo-oracle} \emph{at test time} (it does not create training examples for $\pi_\theta$, thus Line~\ref{line:create_train} is skipped). 
Put differently, the oracle policy selects the set of unlabeled examples that maximizes the target metric of our model on a set sampled from the \emph{same distribution} as the test set.
Therefore, the oracle policy enjoys extremely favorable conditions compared to a trained policy, and we expect it to provide an upper bound on the performance of $\pi_\theta$. Despite these clear advantages, we will show that an oracle policy struggles to substantially improve performance compared to a random policy.

While the oracle policy effectively ``peeks" at the label to make a decision, there are various factors that could explain the low performance of a model trained under the oracle policy. We now list several hypotheses, and in \S\ref{subsec:extended} and \S\ref{sec:analysis} methodologically examine whether they explain the empirical results of LTAL. 
\begin{itemize}[topsep=0pt, itemsep=0pt, leftmargin=0.1in, parsep=0pt]
    \item \textbf{Training}: The models $m^i_{\phi_j}$ are
    affected by the training procedure in Lines~\ref{line:train} and~\ref{line:finetune} of Alg.~\ref{alg:algo-oracle}. Different training procedures affect the performance of models trained with the oracle policy.
    \item \textbf{Search space coverage}: Training over all unlabeled examples in each iteration is intractable, so the oracle policy randomly samples $K$ subsets, each with $L$ examples. Because $K\cdot L << |\unlabset|$, it is possible that randomly sampling these sets will miss the more beneficial unlabeled examples. Moreover, the parameter $L$ controls the diversity of candidate subsets, since as $L$ increases the similarity between the $K$ different subsets grows. Thus, the hyper-parameters $K$ and $L$ might affect the outcome of the oracle policy. 
    \item \textbf{Model architecture}: The model architecture (e.g., number of parameters) can affect the efficacy of learning under the oracle policy.
    \item \textbf{Stochasticity:} 
    \label{subsec:Stochast}
    The oracle policy chooses an unlabeled set based on performance after training with stochastic gradient descent. Differences in performance between candidate sets might be related to this stochasticity, rather than to the actual value of the examples (especially when $\labset$ is small).
    \item \textbf{Myopicity:} The oracle policy chooses the set $\sC^i_j$ that maximizes its performance. However, the success of LTAL depends on the \emph{sequence} of choices that are made. It is possible that the greedy nature of this procedure results in sub-optimal performance. Unfortunately, improving search through beam search or similar measures is intractable in this already computationally-expensive procedure.
\end{itemize}

We now describe QA-SRL~\cite{he2015question}, which is the focus of our investigation, and then describe the experiments with the oracle policy.

\comment{
As in classic Active Learning flow, the \textit{Oracle} policy selects in each iteration $i$ a single candidate $c_{i,j} \subset S_{un}$ that should be annotated and added to the labeled section. Unlike a regular Active Learning acquisition function, the \textit{Oracle} policy is exposed the the ground truth labels of each of the samples in $S_{un}$.

The set of $n$ candidates $\{c_{i,1}, c_{i,2}, \dots c_{i, n}\}$ in each iteration is built by randomly sampling  $l$ samples from $S_{un}$ for each of the $n$ candidates.

Given a candidate $c_{i,j}$ we define its score $s_{i,j}$ as the accuracy of a trained model $m_{i,j}$ on $S_{eval}$, where $m_{i,j}$ was trained using  $S_{lab} \bigcup c_j$. The \textit{Oracle} policy selects the candidate $j$ such that:
\begin{equation}
    j = \arg\max\limits_{j \in 1 \dots n} s_{i,j}
\end{equation}
}

%% file: 04_qasrl_schema.tex
\begin{table*}[t!]
\centering
\small
\begin{tabular}{lll}
\multicolumn{3}{c}{\emph{Elizabeth Warren \textbf{announced} her candidacy at a rally in Massachusetts}. } \\ \toprule
Argument & QA-SRL role & PropBank role \\ \hline \hline
Elizabeth Warren & Who \textbf{announced} something?  & \texttt{ARG0} \\
her candidacy & What did someone \textbf{announce}?   &  \texttt{ARG1} \\
at a rally in Massachusetts & Where did someone \textbf{announce} something? & \texttt{ARGM-LOC} \\\bottomrule
\end{tabular}
\caption{Example of QA-SRL versus traditional SRL annotation for a given input sentence (top). 
Each line shows a single argument, and its role in QA-SRL (in question form) followed by its traditional SRL role, using PropBank notation.
Roles in QA-SRL have a structured open representation, while SRL assigns discrete roles from a predefined set.}
\label{tab:qasrl}
\end{table*}

\section{QA-SRL Schema}
QA-SRL was introduced by \newcite{he2015question} as an open variant of the predefined role schema in traditional SRL.  QA-SRL replaces the predefined set of \emph{roles} 
with the notion of \emph{argument questions}. These are natural language questions centered around a target predicate, where the answers to the given question are its corresponding arguments.
For example, for the 
sentence \emph{``Elizabeth Warren decided to \textbf{run} for president"}, traditional SRL will label \emph{``Elizabeth Warren"} as \texttt{ARG0} of the \textbf{run} predicate (the agent of the predicate, or the entity running in this case), while QA-SRL will assign the more subtle question \emph{``who might run?''}, indicating the uncertainty of this future event. Questions are generated by assigning values to 7 pre-defined slots (where some of the slots are potentially empty). See Table \ref{tab:qasrl} for an example QA-SRL annotation of a full sentence.

Recently, \newcite{fitzgerald2018large} demonstrated the scalability of QA-SRL by crowdsourcing the annotation of a large QA-SRL dataset, dubbed QA-SRL bank 2.0. 
It consists of 250K QA pairs over 64K sentences on three different domains (Wikipedia, news, and science).
Following, this large dataset has enabled the development a neural model which breaks QA-SRL into a pipeline of two tasks, given a target predicate in an input sentence. First, a \emph{span detection} algorithm identifies arguments of the predicate as continuous spans in the sentence (e.g., \emph{``Elizabeth Warren"} in the previous example), then a \emph{question generation} model predicts an appropriate role question (e.g., \emph{``who might run?''}). 

We find that QA-SRL is a good test-bed for active learning of semantic representations, for several key reasons: (1) it requires semantic understanding of the sentence, beyond syntactic or surface-level features (e.g., identifying the factuality of a given predicate), 
(2) adopting the formulation of \newcite{fitzgerald2018large}, it consists of two semantic tasks, allowing us to test active learning on both of them, (3) we can leverage the large QA-SRL dataset to simulate active learning scenarios, and lastly 
(4) QA-SRL's scalability is attractive for the application of active learning policies, as they may further reduce costs for researchers working on developing specialized semantic representations in low-resource domains (e.g., medical, biological, or educational domains).

%% file: 05_basic_exps.tex
\section{Experimental Evaluation}

We now perform a series of experiments comparing the performance of an oracle policy to a random policy. We describe the experimental settings (\S\ref{subsec:exp-settings}), tasks and models (\S\ref{subsec:tasks_and_models}),  present the main results (\S\ref{subsec:basic_results}), and conclude by investigating factors that might affect our empirical findings (\S\ref{subsec:extended}).

\subsection{Experimental Settings}
\label{subsec:exp-settings}
We evaluate the potential of the oracle policy on QA-SRL Bank 2.0~\cite{fitzgerald2018large}. 
We use the training set of the \emph{science} domain as $\sD$, randomly split it into $\labset$, $\unlabset$, and $\evalset$. We evaluate the success of a model $m^i_\phi$ trained with the oracle policy by periodically measuring performance on the development set of the \emph{science} domain. Unless mentioned, all results are an average of 3 experiments, where a different split of $\sD$ was performed. Each experiment used $K$ threads of a 40-core 2.2GHz Xeon Silver 4114 machine.

We compare the results of a base oracle policy  (\bestoracle{}) corresponding to the best policy we were able to obtain using the architecture from~\newcite{fitzgerald2018large} to the following baselines:
\begin{itemize}[topsep=0pt, itemsep=0pt, leftmargin=0.1in, parsep=0pt]
    \item \random: One of the candidate sets $\sC^i_j$ is chosen at random and added to $\labset$.
    \item \longest: The set $\sC^i_j$ with the maximal average number of tokens per sentence is added to $\labset$.
    \item \uncertainty: For each candidate set, we use $m^i_\phi$ to perform predictions over all of the sentences in the set, and choose the set $\sC^i_j$ that has the maximal average entropy over the set of predictions.
\end{itemize}

\subsection{Tasks and Models}
\label{subsec:tasks_and_models}
We now describe the three tasks and corresponding models in our analysis:

\paragraph{Span Detection:} 
Here we detect spans that are arguments of a predicate in a sentence (see Table \ref{tab:qasrl}).
We start with a labeled set of size $|\labset|=50$, and select examples with the oracle policy for $B=460$ iterations. We set the number of candidate sets to $K=5$, and the size of each set to $L=1$, thus the size of the final labeled set is $510$ examples.
We train the publicly available span detection model released by ~\newcite{fitzgerald2018large}, which consumes as input a sentence $x_1, \dots, x_n$, where $x_i$ is the concatenation of the embedding of the $i$th word in the sentence and a learned embedding of a binary indicator for whether this word is the target predicate. This input is fed into a multi-layer encoder, producing a representation $h_i$ for every token. Each span $x_{i:j}$ is represented by concatenating the respective hidden states: $s_{ij} = [h_i; h_j]$.
A fully connected network consumes the span representation $s_{ij}$, and predicts a probability whether the span is an argument or not. 

To accelerate training, we reduce the number of parameters to 488K by freezing the token embeddings, reducing the number of layers in the encoder, and by shrinking the dimension of both the hidden representations and the binary predicate indicator embedding. Following~\newcite{fitzgerald2018large}, we use GLoVe embeddings~\cite{pennington2014glove}.

\paragraph{Question Generation:} 
We generate the question (role) for a given predicate and corresponding argument.
We start with a labeled set of size $|\labset|=500$ and perform $B=250$ iterations, where in each iteration we sample $K=5$ candidate sets each of size $L=10$ (lower values were intractable). Thus, the final size of $\labset$ is 3,000 samples.
We train the publicly available \emph{local} question generation model from~\newcite{fitzgerald2018large}, where the learned argument representation $s_{ij}$ is used to independently predict each of the 7 question slots.
We reduce the number of parameters to 360K with the same modifications as in the span detector model. As a metric for the quality of question generation models, we use its official metric exact match (EM), which reflects the percentage of predicted questions that are identical to the ground truth questions.

\paragraph{Named Entity Recognition:}
To reproduce the experiments of~\newcite{liu2018learning} we run the oracle policy on the CoNLL-2003 NER English dataset \cite{sang2003introduction}, replicating the experimental settings described in~\newcite{liu2018learning} (as their code is not publicly available). We run the oracle policy for $B=200$ iterations, starting from an empty $\labset$, and adding one example ($L=1$) from $K=5$ candidate sets in each iteration.
We use a CRF sequence tagger from AllenNLP \cite{gardner2018allennlp}, and experiment with two variants: (1) \textsc{NER-Multilang}: A Bi-LSTM CRF model (20K parameters) with 40 dimensional multi-lingual word embeddings ~\cite{ammar2016massively}, and (2) \textsc{NER-Linear}: A linear CRF model which was originally used by \newcite{liu2018learning}.

\subsection{Results}
\label{subsec:main_results}
\label{subsec:basic_results}
\paragraph{Span Detection:}
Table~\ref{tab:span-acc--tb} shows $F_1$ score (the official metric) of the QA-SRL span detector models for different sizes of $\labset$ for \bestoracle{} and the other baselines. Figure~\ref{fig:results} (left) shows the relative improvement of the baselines over \random{}. We observe that the maximal improvement of \bestoracle{} over \random{} is 9\% given 200 examples, but with larger $\labset$ the improvement drops to less than 5\%. This is substantially less than the improvements obtained by ~\newcite{liu2018learning} on text classification and NER. 
Moreover, \longest{} outperforms
\bestoracle{} in most of the observed results. This shows that there exists a selection strategy that is better than \bestoracle{}, but it is not the one chosen by the oracle policy.

\begin{table*}[t]
 \center
 {\footnotesize
\begin{tabular}{|lccccccccc|}
\hline \# samples & 100 & 150 & 200 & 250 & 300 & 350 & 400 & 450 & 500 \\ \hline \hline
\bestoracle{} & 42.7& \textbf{49.2}& 52.9& 54.2& \textbf{56.6}& \textbf{57.4}& 58.4& \textbf{59.5}& 59.9 \\ 
\random{} & 42.8& 47.2& 48.3& 52.4& 53.3& 56.1& 57.0& 57.5& 58.5  \\ 
\longest & \textbf{44.1}& 49.1& \textbf{53.0}& \textbf{55.5}& 56.4& \textbf{57.4}& \textbf{58.7}& 58.6& \textbf{60.0} \\ 
\uncertainty & 42.8& 47.0& 50.1& 51.3& 52.2& 54.4& 55.1& 55.6& 56.9 \\ \bottomrule
\end{tabular}
}
\caption{Span detection $F_1$ on the development set for all models across different numbers of labeled examples.}
\label{tab:span-acc--tb}

\end{table*}

\input{99-oracle-random-diff-plot}

\paragraph{Question Generation:}
To check whether the previous result is specific to span detection, we conduct the same experiment for question generation. However, training question generation models is slower compared to span detection and thus we explore a smaller space of hyper-parameters. 
Table~\ref{tab:question-acc--tb} reports the EM scores achieved by \bestoracle{} and the other baselines, and Figure~\ref{fig:results} (center) shows the relative improvement. Here, the performance of \bestoracle{} is even worse compared to span detection, as its maximal relative improvement over \random{} is at most 5\%.

\begin{table*}[t]
\center
{\footnotesize
\begin{tabular}{|lccccccccc|}
\hline \# samples & 550 & 750 & 1000 & 1250 & 1500 & 1800 & 2100 & 2500 & 3000 \\ \hline \hline
\bestoracle{} & 18.9& \textbf{21.7}& \textbf{24.4}& \textbf{26.4}& 27.1& \textbf{28.4}& \textbf{29.1}& \textbf{30.6}& 31.1 \\ 
\random{} & 18.1& 21.4& 23.7& 25.2& \textbf{27.3}& 27.9& 28.6& 29.9& 31.3  \\ 
\longest{} & 17.8& 20.9& 22.8& 25.7& 27.1& 28.0& \textbf{29.1}& 30.4& 31.3 \\
\uncertainty & \textbf{19.3}& 21.4& 23.0& 25.4& 26.9& 27.8& 29.0& 29.8& \textbf{31.5} \\ \bottomrule

\end{tabular}
\caption{Question generation  scores (exact match) on the development set across different numbers of labeled examples.
}
\label{tab:question-acc--tb}
}
\end{table*}

\noindent
\paragraph{Named Entity Recognition:}
Figure~\ref{fig:results} (right) shows the relative improvement of \textsc{NER-Linear} and \textsc{NER-MultiLang} compared to \random{}. We observe that in \textsc{NER-Linear}, which is a replication of~\newcite{liu2018learning}, the oracle policy indeed obtains a large improvement over \random{} for various sizes of $\labset$, with at least 9.5\% relative improvement in performance. However, in \textsc{NER-MultiLang} the relative gains are smaller, especially when the size of $\labset$ is small.

%% file: 99-oracle-random-diff-plot.tex
 
  
\begin{figure*}[htbp]
\centering
\begin{tikzpicture}
\begin{groupplot}[group style={group size=3 by 1}, width=0.365\textwidth]
    \nextgroupplot[
    title={QA-SRL span detection},
    xlabel={Size of $\labset$},
    ylabel={},
    xmin=100, xmax=550,
    ymin=-5, ymax=25,
    xtick={100,200,300,400,500},
    ytick={0,5,10,15,20},
    legend pos=north west,
    ymajorgrids=true,
    grid style=dashed]
 
    \addplot[ color=blue, mark=square]
   coordinates {(110,-1.76)(150,4.32)(210,4.77)(290,7.03)(370,3.69)(510,3.09)};
   \addlegendentry{\small \bestoracle{}}
   \addplot[color=green,mark=square]
   coordinates {(110,0.34)(150,2.43)(210,3.99)(290,4.63)(370,4.01)(510,2.27)};
   \addlegendentry{\small \textsc{L=5}}
   \addplot[color=red,mark=square]
   coordinates {(110,3.52)(150,3.34)(210,2.96)(290,1.21)(370,0.13)(510,1.14)};\addlegendentry{\small \textsc{OracleSmallModel}}
   
    \nextgroupplot[
    title={QA-SRL question generation},
    xlabel={Size of $\labset$},
    ylabel={},
    xmin=500, xmax=3100,
    ymin=-10, ymax=25,
    xtick={500,1500,2500},
    ytick={0, 5,10,15, 20},
    legend pos=north west,
    ymajorgrids=true,
    grid style=dashed]
    \addplot[ color=blue, mark=square]
  coordinates {(550,4.49)(750,1.62)(1000,2.69)(1250,4.86)(1500,-0.76)(1800,1.95)(2100,1.79)(2500,2.16)(3000,-0.45)};
   \addlegendentry{\small \textsc{BaseOracle}}
   \addplot[ color=green, mark=square]
   coordinates {(550,-1.92)(750,-2.16)(1000,-4.02)(1250,2.17)(1500,-0.77)(1800,0.58)(2100,1.91)(2500,1.69)(3000,0.06)};
   \addlegendentry{\small \textsc{Longest}}
   \addplot[ color=red, mark=square]
   coordinates {(550,6.83)(750,0.19)(1000,-3.31)(1250,1.05)(1500,-1.60)(1800,-0.18)(2100,1.59)(2500,-0.50)(3000,0.63)};
   \addlegendentry{\small \textsc{Uncertainty}}
   
   \nextgroupplot[
    title={NER},
    xlabel={Size of $\labset$},
    ylabel={},
    xmin=30, xmax=220,
    ymin=-5, ymax=90,
    xtick={50,100,150,200},
    ytick={0, 10, 20, 40, 60},
    legend pos=north east,
    ymajorgrids=true,
    grid style=dashed]
 
   \addplot[color=blue,mark=square]
   coordinates {(40,18.30)(60,20.08)(80,14.45)(100,11.58)(120,9.81)(140,6.86)(160,7.20)(180,7.13)(200,7.13)};
   \addlegendentry{\small \textsc{MultiLang}}
   \addplot[color=green, mark=square]
   coordinates {(40,85.04)(60,50.46)(80,33.47)(100,19.31)(120,15.46)(140,13.04)(160,10.77)(180,11.13)(200,9.67)};
    \addlegendentry{\small \textsc{Linear}}

\end{groupplot}
\end{tikzpicture}
\caption{Relative improvement (in \%) of different models compared to \random{} on the development set. Note that the range of the y-axis in NER is different from QA-SRL.}
\label{fig:results}
\end{figure*}
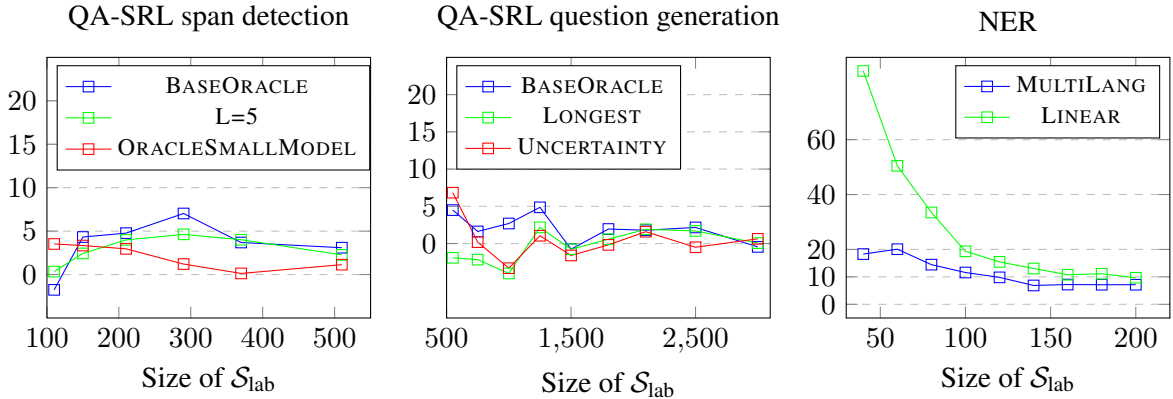

%% file: 06_extended_exsps.tex
\subsection{Extended Experiments}

\begin{table}[t]
  \center
  \resizebox{\columnwidth}{!}{ 
\begin{tabular}{|lcccccc|}
\hline \# samples & 110 & 150 & 210 & 290 & 370 & 510 \\ \hline \hline
  \random{}  & 45.2& 47.2& 50.5& 53.1& 55.8& 58.5  \\ 
 \bestoracle{} & 43.3& \textbf{49.2}& \textbf{52.9}& \textbf{56.8}& 57.8& \textbf{60.3} \\ \hline \hline
\textsc{$K=10$} & \textbf{46.6}& 48.8& 51.4& 55.8& 57.6& 58.6 \\ 
\textsc{$K=20$} & 44.8   & 47.4   & 52.1   & 55.9   & ---   & --- \\ 
\textsc{$L=5$} & 44.2& 48.3& 52.5& 55.5& \textbf{58.0}& 59.8 \\ 
\textsc{$L=20$} &45.2& 47.5& 51.9& 55.1& 56.7& 58.7  \\ \hline \hline
\textsc{Loss-Score} & 30.0& 38.2& 41.0& 51.5& 53.7& 57.2  \\ 
\textsc{Indep.}{*}  & 40.9& 44.6& 50.1& 54.1  & ---  & ---  \\ \hline \hline
\textsc{Epsilon-Greedy-0.3} & 45.1& 48.6& 52.4& 55.6& 57.1& 59.8 \\ 
\textsc{Oracle-100} & 44.9& 48.8& 51.4& 53.9& 57.0& 59.2 \\ \hline \hline
\textsc{RandomSmallModel} & 51.9& 54.8& 57.3& 59.5& 61.4& 62.6 \\ 
\textsc{OracleSmallModel} & 53.8& 56.6& 58.9& 60.3& 61.5& 63.3  \\ \bottomrule
\end{tabular}
}
\caption{Span detection $F_1$ scores on the development set for different size of $\labset$. We highlight the best performing policy for the standard span detector architecture. (\text{*}) indicates that the results are for a single run.}
\label{tab:span-acc-extend-tb}
  
\end{table}

\label{subsec:extended}
Surprisingly, we observed in \S\ref{subsec:main_results} that even an oracle policy, which is allowed to pick the examples that maximize performance on samples from the same distribution as the test set, does not substantially improve performance over a random policy.
One possibility is that no active learning policy is better than random. However, \longest{} outperformed \bestoracle{} showing that the problem is at least partially related to \bestoracle{} itself.

We now examine the possible factors described in \S\ref{sec:oracle} and
investigate their interaction with the performance of models trained with \bestoracle{}.
All modifications were tested on \textit{span detection}, using the
experimental settings described in \S\ref{subsec:exp-settings}.

\paragraph{Search space coverage} We begin by examining the effect of the
parameters $K$ and $L$ on the oracle policy. As $K$ increases, we cover more of the unlabeled data, but training time increases linearly. 
As $L$ increases, the subsets $\{\sC_j\}_{j=1}^K$ become more similar to one another due to the fact that we are randomly mixing more examples from the unlabeled data. On the other hand, when $L$ is small, the fine-tuning process is less affected by the candidate sets and more by $\labset$. In such case, it is likely that the difference in scores is also affected by stochasticity.

\bestoracle{} uses $K=5, L=1$. We examine the performance of the oracle policy as
these values are increased in Table~\ref{tab:span-acc-extend-tb}. We observe
that performance does not improve, and perhaps even decreases for larger values
of $K$. We hypothesize that a large $K$ increases the greediness of the procedure,
and may result in selecting an example that seems promising in the current
iteration but is sub-optimal in the long run, similar to large beam sizes
reducing performance in neural machine translation \cite{yang2018breaking}. A
moderate $K$ results in a more random and possibly beneficial selection.

Increasing the size of each candidate set to $L=5$ or  $20$ results in roughly similar
performance to $L=1$. We hypothesize that there is a trade-off where as $L$ increases
the similarity between the different sets increases but training becomes more
stable and vice versa, and thus performance for different $L$ values does not vary substantially.

\paragraph{Training}
In Lines~\ref{line:train} and~\ref{line:finetune} of Alg.~\ref{alg:algo-oracle} we train on $\labset$ and then
fine-tune on the union $\labset \cup \sC^i_j$ until $s^i_j$ does not significantly improve for 5 epochs. It
is possible that fine-tuning from a fixed model reduces the efficacy of
training, and training on $\labset \cup \sC^i_j$ from random weights will improve
performance. Of course, training from scratch will substantially increase
training time. We run an experiment, termed \textsc{Indep.}, where
Line~\ref{line:train} is skipped, and in Line~\ref{line:finetune} we
independently train each of the candidate models from random weights. We find
that this modification does not achieve better results than \bestoracle{}, possibly because training a model from scratch for each of the candidates increases the stochasticity in the optimization. 

In addition, we also experiment with fine-tuning on $\sC_j$ only, rather than $\labset \cup \sC_j$. As we expect, results are quite poor since the model uses only a few examples for fine-tuning and forgets the examples in the labeled set. 

Lastly, we hypothesize that selecting a candidate set based on the target metric ($F_1$ for span detection) might not be sensitive enough and thus we run an experiment, termed \textsc{Loss-Score}, where we select the set $\sC_j$ that minimizes the loss on the development set. 
We find that this modification achieves lower results than \random{}, especially when $\labset$ is small,  reflecting the fact that the loss is not perfectly correlated with our target metric.

\paragraph{Model Architecture} 
In \S\ref{subsec:main_results} we observed that results on NER vary with the model architecture. To see whether this phenomenon occurs also for span detection we perform a modification to the model -- we reduce the number of parameters from 488K to 26K by reducing the hidden state size and replacing GLoVe embeddings with multi-lingual embeddings~\cite{ammar2016massively}. We then compare an oracle policy (\textsc{OracleSmallModel}) with a random policy (\textsc{RandomSmallModel}). Table~\ref{tab:span-acc-extend-tb} shows that while absolute $F_1$ actually improves in this setup, the oracle policy improves performance compared to a random policy by no more than 4\%. Thus, contrary to NER, here architecture modifications do not expose an advantage of the
oracle policy compared to the random one. We did not examine a simpler linear model for span detection, 
in light of recent findings~\cite{Lowell2018PracticalOT} that it is important to test LTAL with state-of-the-art models, as performance is tied to the specific model being trained.

\paragraph{Myopicity}
We hypothesized that greedily selecting an example that maximizes performance in a specific iteration might be suboptimal in the long run. Because  non-greedy selection strategies are computationaly intractable, we perform the following two experiments. 

First, we examine \textsc{Epsilon-Greedy-p}, where in each iteration the oracle policy selects the set $\sC_j$ that maximizes target performance with probability $1-p$ and randomly chooses a set with probability $p$. This is meant to check whether adding random exploration to the oracle policy might prevent it from getting stuck in local optima. We find that when $p=0.3$ its performance is comparable to \bestoracle{} while reducing the computational costs.

Second, we observe that most of the gain of \bestoracle{} compared to \random{} is in the beginning of the procedure. Thus, we propose to use \bestoracle{} in the first $b$ iterations, and then transition to a random policy (termed \textsc{Oracle-b}). We run this variation with $b=100$ and find that it leads to similar performance.

To summarize, we have found that an oracle policy only slightly improves performance for QA-SRL span detection and question generation compared to a random policy, and that improvements in NER are also conditioned on the underlying model. Our results echo recent findings by~\newcite{Lowell2018PracticalOT}, who have shown that gains achieved by active learning are small and inconsistent when modifying the model architecture. 
 
We have examined multiple factors that might affect the performance of models trained with an oracle policy including the training procedure, model architecture, and search procedure, and have shown that in all of them the oracle policy struggles to improve over the random one. Thus, a \emph{learned policy} is even less likely to obtain meaningful gains using LTAL.
 
In the next section we analyze the differences between \textsc{NER-LINEAR}, where LTAL works well, and \bestoracle{}, in order to better understand the underlying causes for this phenomenon.

%% file: 07_analysis.tex
\section{When does LTAL Work?}
\label{sec:analysis}
\input{98_analysis_fig}
A basic underlying assumption of active learning (with or without a learned policy), is that some samples in $\unlabset$ are more informative for the learning process than others. In LTAL, the informativeness of a candidate example set is defined by the accuracy of a trained model, as evaluated on $\evalset$ (Line \ref{line:score} in Alg.~\ref{alg:algo-oracle}). Thus, for active learning to work, the candidate set that is selected should not be affected by the stochasticity of the training process.
Put differently, the ranking of the candidate sets by the oracle policy should be \emph{consistent} and not be dramatically affected by the optimization.

To operationalize this intuition, we use Alg.~\ref{alg:algo-oracle}, but run the for-loop in Line~\ref{line:for} twice, using two different random seeds. 
Let $\sC_t^i$ be the chosen or \emph{reference} candidate set according to the first run of the for-loop in iteration $i$. We can measure the consistency of the optimization process by looking at the ranking of the candidate sets $\sC_1^i, \dots \sC_K^i$ according to the second fine-tuning, and computing the mean reciprocal rank (MRR) with respect to the reference candidate set $\sC_t^i$ across all iterations:
\begin{equation}
    \textnormal{MRR} = \frac{1}{B} \sum_{i=1}^{B} \frac{1}{\text{rank}(\sC^i_t)},
\end{equation}
where  $\text{rank}(\sC^i_t)$ is the rank of $\sC^i_t$ in the second fine tuning step. 
The only difference between the two fine-tuning procedures is the random seed. Therefore, an MRR value that is close to $1$ means that the ranking of the candidates is mostly affected by the quality of the samples, while a small MRR hints that optimization plays a large role. 
We prefer MRR to other correlation-based measures (such as Spearman's rank-order correlation), because the oracle is only affected by the candidate set that is ranked first.
We can now examine whether the MRR score correlates with whether LTAL works or not.

We measure the MRR in 3 settings: (1) \textsc{Ner-Linear}, a linear CRF model for NER which replicates the experimental settings in~\cite{liu2018learning}, where LTAL works, (2) \textsc{Ner-Multilang}, a BiLSTM-CRF sequence tagger from AllenNLP~\cite{gardner2018allennlp} with 40 dimensional multi-lingual word embeddings of~\newcite{ammar2016massively}, and (3) \bestoracle{}, the baseline model for span detection task. In all experiments the initial $\labset$ was empty and $B=200$, following the experimental settings in which LTAL has shown good performance \cite{liu2018learning, fang2017learning, vu2019learning}.  Since the MRR might change
as the size of $\labset$ is increasing, we compute and report MRR every 10 iterations.

Figure~\ref{fig:analysis} (left) presents the MRR in the three experiments. We observe that in \textsc{NER-Linear} the MRR has a stable value of $1$, while in \textsc{Ner-Multilang} and \bestoracle{} the MRR value is substantially lower, and closer to an MRR value of a random selection ($\sim$.46). The right side of Figure~\ref{fig:analysis} shows that \textsc{NER-Linear} oracle policy outperforms a random policy by a much larger margin, compared to the other 2 experiments.

These results show that the ranking in \textsc{NER-Linear} is not affected by the stochasticity of optimization, which is expected given its underlying convex loss function. On the other hand, the optimization process in the other experiments is over a non-convex loss function and a small $\labset$, and thus optimization is more brittle. 
Interestingly, we observe in Figure~\ref{fig:analysis} that the gains of the oracle policy in \textsc{NER-Linear} are higher than \textsc{NER-Multilang}, although the task and the dataset are exactly same in the two experiments. 
This shows that the potential of LTAL is affected by the model, where a more complex model leads to smaller gains by LTAL.


We view our findings as a guideline for future work: by tracking the MRR one can assess the potential of LTAL at development time -- when the MRR is small, the potential is limited.


%% file: 98_analysis_fig.tex
\begin{figure*}[htbp]
\centering
\begin{tikzpicture}
    \begin{groupplot}[group style={group size=2 by 1}, width=0.37\textwidth]
    \nextgroupplot[
    title={MRR},
    xlabel={Size of $\labset$},
    ylabel={},
    xmin=0, xmax=205,
    ymin=0, ymax=1.2,
    xtick={50, 100 , 150, 200},
    ytick={0, 0.2, 0.4, 0.6, 0.8, 1},
    ymajorgrids=true,
    grid style=dashed]
 
  \addplot[color=green,mark=square]
   coordinates {(10,1.00)(20,1.00)(30,1.00)(40,1.00)(50,1.00)(60,1.00)(70,1.00)(80,1.00)(90,1.00)(100,1.00)(110,1.00)(120,1.00)(130,1.00)(140,1.00)(150,1.00)(160,1.00)(170,1.00)(180,1.00)(190,1.00)(200,1.00)};
  \addplot[color=red,mark=square]
  coordinates {(10,0.78)(20,0.57)(30,0.52)(40,0.58)(50,0.45)(60,0.46)(70,0.55)(80,0.29)(90,0.62)(100,0.41)(110,0.44)(120,0.37)(130,0.38)(140,0.42)(150,0.37)(160,0.62)(170,0.47)(180,0.39)(190,0.59)(200,0.28)};
  \addplot[ color=blue, mark=square]
  coordinates {(10,0.69)(20,0.55)(30,0.37)(40,0.35)(50,0.45)(60,0.40)(70,0.52)(80,0.37)(90,0.50)(100,0.41)(110,0.40)(120,0.47)(130,0.32)(140,0.54)(150,0.50)(160,0.58)(170,0.47)(180,0.43)(190,0.49)(200,0.65)};

    \nextgroupplot[
    title={Relative improvement (in \%)},
    xlabel={Size of $\labset$},
    ylabel={},
    xmin=30, xmax=205,
    ymin=-10, ymax=90,
    xtick={50, 100 , 150, 200},
    ytick={0, 20, 40, 60, 80},
    ymajorgrids=true,
    grid style=dashed,
    legend columns=3,
    legend to name=grouplegend]
    
     \addplot[color=green, mark=square]
    coordinates {(40,85.04)(60,50.46)(80,33.47)(100,19.31)(120,15.46)(140,13.04)(160,10.77)(180,11.13)(200,9.67)};
    \addlegendentry{\small \textsc{NER-Linear}}
    \addplot[color=red,mark=square]
   coordinates {(40,18.30)(60,20.08)(80,14.45)(100,11.58)(120,9.81)(140,6.86)(160,7.20)(180,7.13)(200,7.13)};
    \addlegendentry{\small \textsc{NER-Multilang}}
    \addplot[color=blue, mark=square] 
   coordinates {(40,48.53)(60,16.92)(80,9.29)(100,7.57)(140,1.52)(180,1.09)(200,2.54)};\addlegendentry{\small \bestoracle}

    \end{groupplot}
    
    \node[below=1em] at(current bounding box.south) {\pgfplotslegendfromname{grouplegend}};
\end{tikzpicture}
\caption{ MRR (on the left) and relative improvement (in \%) of different models compared to \random{} on the development set.}
\label{fig:analysis}
\end{figure*}
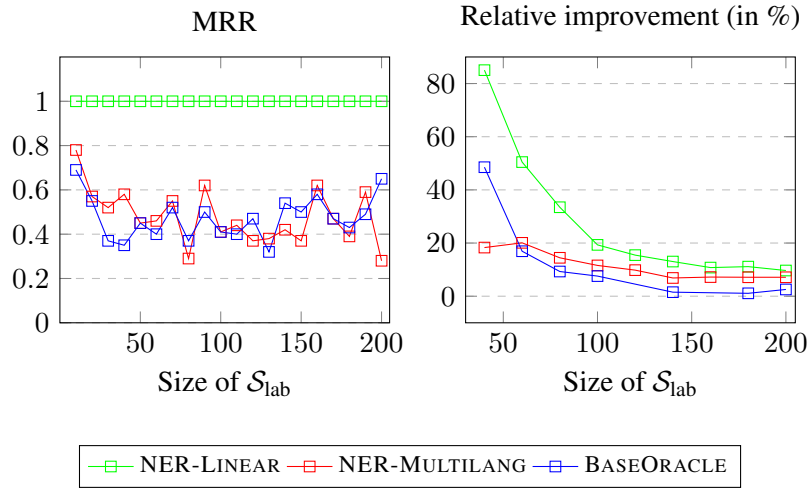

\comment{\begin{figure*}[tbh]
\begin{center}
\begin{subfigure}{.33\textwidth}
  \includegraphics[width=\linewidth]{resources/NER_MULTILANG_MODEL}
  \caption{\textsc{Ner-Multilnag}}
  \label{fig:sub1}
\end{subfigure}%
\begin{subfigure}{.33\textwidth}
  \includegraphics[width=\linewidth]{resources/SPAN_GLOVE_MODEL}
  \caption{\textsc{base-oracle}}
  \label{fig:sub1}
\end{subfigure}%
\begin{subfigure}{.33\textwidth}
  \includegraphics[width=\linewidth]{resources/SPAN_MULTILANG_MODEL.png}
  \caption{\textsc{OracleSmallModel}}
  \label{fig:sub2}
\end{subfigure}
\caption{\textsc{SD-Between} (right y-axis) and relative improvement (in \%, left y-axis) over \random\  as function of $|\labset|$.}
\label{fig:analysis}
\end{center}
\end{figure*}}

%% file: 08_related.tex
\section{Related Work}
\label{sec:related}

Active learning has shown promising results on various tasks. The commonly used \textit{uncertainty} criteria ~\cite{lewis1994heterogeneous, culotta2005reducing} is focused on selecting the samples on which the confidence of the model is low. Among other notable approaches, in \textit{query by committee} ~\cite{seung1992query} a disagreement
between a set of trained models on the predicted output of an unlabeled sample is the criterion for selecting what samples to label. 

In a large empirical study, \newcite{Lowell2018PracticalOT} have recently shown other limitations in active learning. They investigate the performance of active learning across NLP tasks and model architectures, and demonstrate that it does not achieve consistent gains over supervised learning, mostly because the collected samples are beneficial to a specific model architecture, and does not yield better results than random selection when switching to a new architecture.

There has been little research regarding active learning of semantic representations. Among the relevant work, \newcite{siddhant2018deep} have shown that \textit{uncertainty} estimation using dropout and Bayes-By-Backprop \cite{blundell2015weight} achieves good results on the SRL formulation. The improvements in performance due to LTAL approaches on various tasks ~\cite{konyushkova2017learning, bachman2017learning, fang2017learning, liu2018learning} has raised the question whether learned policies can be applied also to the field of learning semantic representations.


%% file: 09_conclusions.tex
\section{Conclusions}
\label{sec:conclusions}

We presented the first experimentation with LTAL techniques in learning parsers for semantic representations. 
Surprisingly, we find that LTAL, a learned method which was shown to be effective for NER and document classification, does not do significantly better than a random selection on two semantic representation tasks within the QA-SRL framework, even when given extremely favourable conditions.
We thoroughly analyze the factors leading to this poor performance, and find that the stochasticity in the model optimization negatively affects the performance of LTAL. 
Finally, we propose a metric which can serve as an indicator for whether LTAL will fare well for a given dataset and model. Our results suggest that different approaches should be explored for the important task of building semantic representation models.

%% file: 11_ackno.tex
\section*{Acknowledgements}
We thank Julian Michael and Oz Anani for their useful comments and feedback.
This research was supported by The U.S-Israel Binational Science Foundation grant 2016257,
its associated NSF grant 1737230 and
The Yandex Initiative for Machine Learning.